%% file: root.tex
\documentclass[letterpaper, 10 pt, conference]{ieeeconf}  

\IEEEoverridecommandlockouts                              

\usepackage{cuted}
\usepackage{tabularx}
\usepackage{gensymb}
\usepackage{multirow}
\usepackage{adjustbox}
\usepackage{array}
\usepackage{booktabs}
\usepackage{commath}
\usepackage{caption}
\usepackage[dvipsnames]{xcolor}
\usepackage{hyperref}

\usepackage{listings}
\lstdefinestyle{jsonbw}{
    language=,
    basicstyle=\small\ttfamily,
    breaklines=true,
    frame=lines,
    showstringspaces=false,
    emph={
        "model", "api_call", "payload", "instructions", "input",
        "role", "content", "type", "text", "image_url", "user",
        context_text, b64_img, b64_img1, b64_img2, category,
        object_name, scheme, dataset, min_frame_gap
    },
    emphstyle=\bfseries,
}

\title{\LARGE \bf
Size Matters: Reconstructing Real-Scale 3D Models from \\ Monocular Images for Food Portion Estimation
}

\author{
Gautham Vinod$^*$\quad
Bruce Coburn$^*$\quad
Siddeshwar Raghavan$^*$\quad
{Jiangpeng He}$^{\dagger}$ \quad
Fengqing Zhu$^*$
\\
\small {$^*$ Purdue University, West Lafayette, Indiana, U.S.A. \quad \quad $^{\dagger}$ Indiana University, Bloomington, Indiana, U.S.A}  \\
{\tt\small \{gvinod, coburn6, raghav12, zhu0\}@purdue.edu \quad \{jhe2\}@iu.edu} }

\usepackage[compact]{titlesec}
\titlespacing{\section}{0pt}{2ex}{1ex}
\titlespacing{\subsection}{0pt}{1ex}{0.5ex}
\titlespacing{\subsubsection}{0pt}{0.5ex}{0ex}

\usepackage[font=small,labelfont=bf,skip=2pt]{caption}

\begin{document}

\maketitle
\begin{strip}
\begin{minipage}{\textwidth}\centering
\vspace{-1cm}
\includegraphics[width=0.9\textwidth]{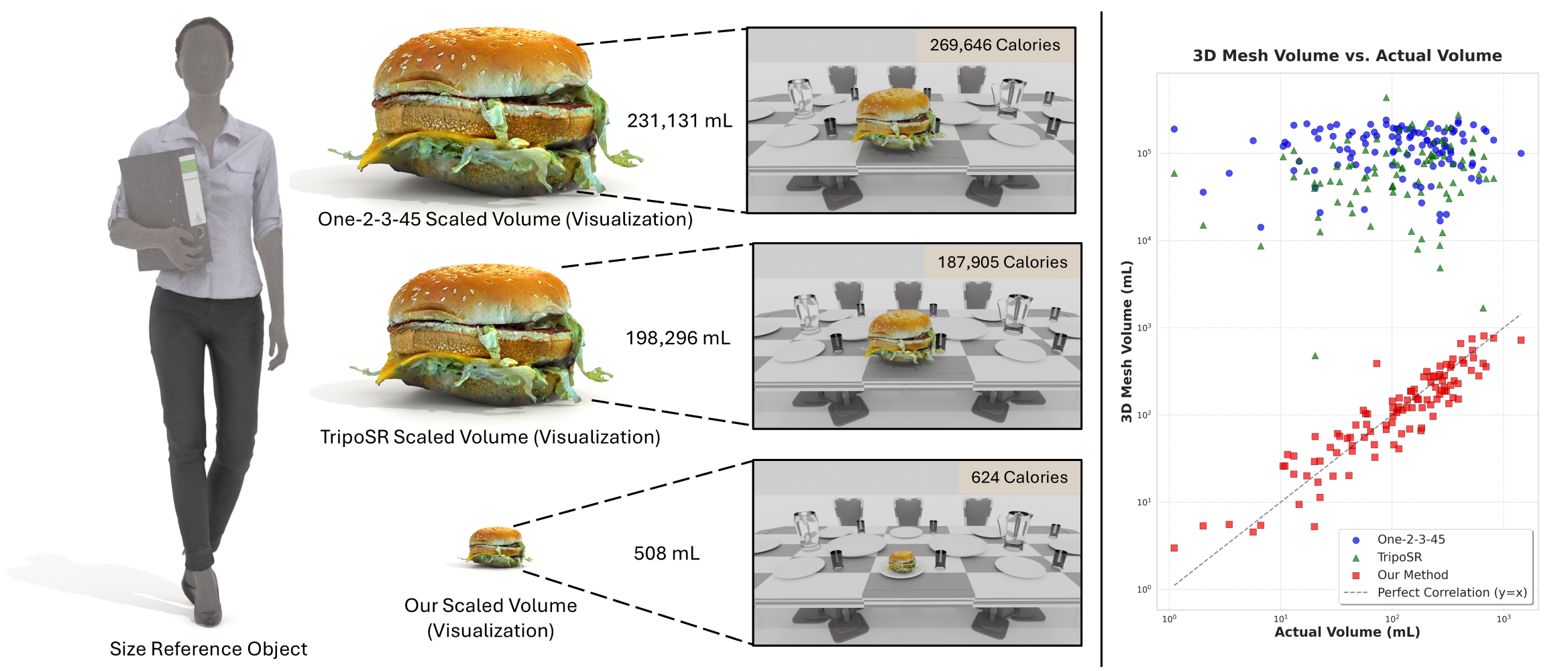}
\captionof{figure}{\textbf{Restoring real-world scale in single-view 3D reconstruction.} Existing single-view reconstruction methods, such as One-2-3-45~\cite{liu2023one-2-3-45} and TripoSR~\cite{tochilkin2024triposr}, generate 3D meshes with arbitrary scales leading to unrealistic size interpretation.  In contrast, our approach accurately produces 3D objects at a real-world scale, closely matching actual volumes. For visualization, rather than using the actual reconstructed 3D models, we rescale the original 3D model to match the volumes produced by each method. The visualization (left) highlights the disparity (actual size of the reconstruction compared to a human as a size reference) in physical scale, demonstrating the need for accurate real-scale reconstruction, while the scatter plot (right) demonstrates our method’s significantly improved alignment with ground-truth measurements.}
\label{fig:intro}
\end{minipage}
\end{strip}
\vspace{-1cm}
\thispagestyle{empty}
\pagestyle{empty}


\input{sections/0_abstract}
\input{sections/1_intro}
\input{sections/2_related_works}
\input{sections/3_method}
\input{sections/4_experimental_results}
\input{sections/5_conclusion}

{
\small
\bibliographystyle{IEEEtran}
\bibliography{references}
}

\end{document}

%% file: sections/0_abstract.tex
\begin{abstract}

The rise of chronic diseases related to diet, such as obesity and diabetes, emphasizes the need for accurate monitoring of food intake. While AI-driven dietary assessment has made strides in recent years, the ill-posed nature of recovering size (portion) information from monocular images for accurate estimation of ``how much did you eat?'' is a pressing challenge. Some 3D reconstruction methods have achieved impressive geometric reconstruction but fail to recover the crucial real-world scale of the reconstructed object, limiting its usage in precision nutrition. In this paper, we bridge the gap between 3D computer vision and digital health by proposing a method that recovers a true-to-scale 3D reconstructed object from a monocular image. Our approach leverages rich visual features extracted from models trained on large-scale datasets to estimate the scale of the reconstructed object. This learned scale enables us to convert single-view 3D reconstructions into true-to-life, physically meaningful models. Extensive experiments and ablation studies on two publicly available datasets show that our method consistently outperforms existing techniques, achieving nearly a 30\% reduction in mean absolute volume-estimation error, showcasing its potential to enhance the domain of precision nutrition. \textcolor{purple}{Code: \href{https://gitlab.com/viper-purdue/size-matters}{https://gitlab.com/viper-purdue/size-matters}}
\end{abstract}

%% file: sections/1_intro.tex
\section{Introduction}
\label{sec:intro}

The escalation of obesity, diabetes, and cardiovascular disorders drives a global health crisis~\cite{wang2023optimal}, creating an urgent need for more sophisticated preventative strategies. Precision Nutrition, which is the ability of tailoring dietary recommendations based on accurate monitoring of food intake~\cite{he2024MetafoodChallenge}, offers a pivotal mechanism in the prevention of these conditions. While recent AI-driven tools have advanced image-based dietary assessment, they remain fundamentally limited by their inability to extract precise portion (size) information from monocular images. This deficiency stems from a lack of 3D geometric understanding~\cite{Dehais2017TwoViewFoodReconstruction, he2024MetafoodChallenge}, a capability that is indispensable not only for nutritional analysis but also for applications ranging from medical imaging to robotics~\cite{medical_image_processing, liu2005system, e-commerce3d, remondino20143d, robotics_3d, autonomous_3d, manufacturing_3d}.


Recent developments in 3D computer vision such as Neural Radiance Fields (NeRF)~\cite{mildenhall2020nerf} and single-view diffusion models~\cite{liu2023one-2-3-45, tochilkin2024triposr}, have demonstrated remarkable efficacy in reconstructing geometry from single images. However, they suffer from a fundamental limitation for many applications: they lack real-world physical units of scale. These generative models reconstruct objects in an arbitrary canonical space, where a blueberry and a pumpkin may appear to be the same size. Consequently, the full potential of monocular 3D reconstruction cannot be realized without accurate scale information of the 3D model.

In this paper, we bridge the gap between generative 3D vision and precision healthcare. We introduce a method capable of reconstructing 3D objects while simultaneously recovering their real-world scale from monocular images, thereby facilitating accurate portion estimation. Our approach leverages image encoders, specifically CLIP~\cite{radford2021learning}, to extract rich visual features that encode implicit scene and scale information. By integrating these features with single-view 3D reconstruction methods, we train a model to predict a scale factor defined as the ratio between the reconstructed model's volume and the real object's volume. This enables precise scaling of the 3D model to match its true size.

A key strategy in our framework is to incorporate both the input image and multi-view rendering of the reconstructed model. By analyzing the reconstructed object from different perspectives, our model learns to infer and correct scale discrepancies, ensuring accurate real-world size predictions. Finally, the estimated scale factor is applied to the 3D model, and its final volume is compared against the original object for evaluation. As illustrated in Figure~\ref{fig:intro}, previous methods fail to capture real-world scale, limiting their usability in practical applications. 

Our method is evaluated on the MetaFood3D~\cite{chen2024metafood3d} and OmniObject3D~\cite{wu2023omniobject3d} datasets. Through our results, we demonstrate that our method reduces volume estimation error by nearly 30\% compared to existing baselines. Furthermore, we convert these volumetric estimates into caloric energy values, showing that our real-scale reconstruction framework significantly outperforms current methods in estimating nutrition values. Our method provides an innovative solution to image-based dietary assessment, unlocking new possibilities for AI-assisted personal health monitoring. While our primary application is precision nutrition, we validate the generalizability of our framework on the OmniObject3D dataset~\cite{wu2023omniobject3d}, demonstrating that our scale recovery method is effective for generic tasks beyond food items.

The main contributions of our paper are:
\begin{itemize}
    \item We introduce a real-scale reconstruction method that integrates visual priors with geometric inference for scale-aware 3D reconstruction from monocular images.
    \item We propose a volume-based scale predictor that uses shape and semantic features to estimate the correct physical scale of reconstructed objects.
    \item We demonstrate that our approach surpasses existing scale estimation methods, achieving an improvement of approximately 30\% in mean absolute volume error.
\end{itemize}

%% file: sections/2_related_works.tex
\section{Related Works}
\label{sec:related_works}
\begin{figure*}[ht!]
    \centering
    \includegraphics[width=0.9\linewidth]{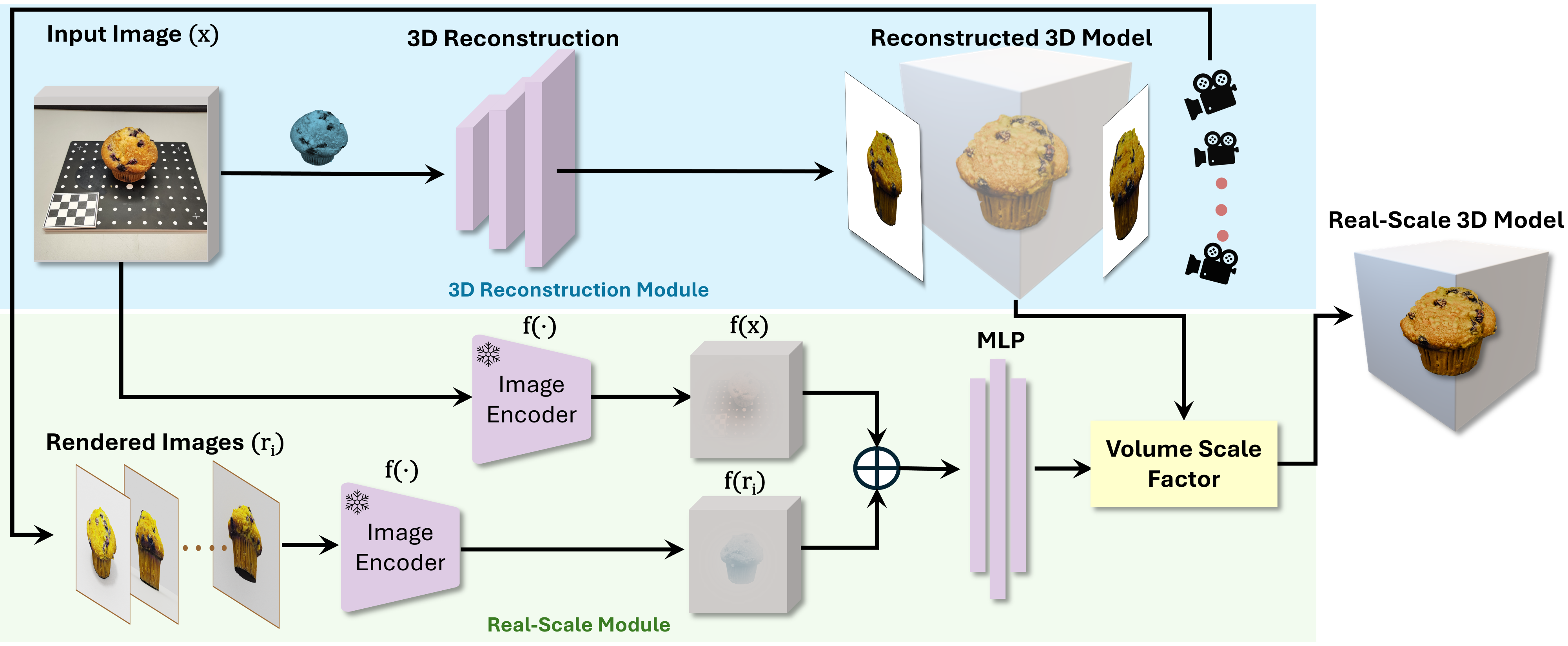}
    \caption{\textbf{Method Overview.} The 3D reconstruction module relies on a single input image and outputs a 3D model using a single-view 3D reconstruction framework. The reconstructed 3D model is used to produce multiple image renders of this model from different viewpoints. Each of these viewpoints, along with the input image, is passed to the Real-Scale Module, where the features are extracted, combined, and put through a network. This network learns the volume scale factor to rescale the reconstructed 3D model to real physical dimensions.}
    \vspace{-0.5cm}
    \label{fig:method_overview}
\end{figure*}
\subsection{3D Reconstruction}
3D Reconstruction has relied on representations such as voxels~\cite{Shao2023VoxelReconstruction, xie2019pix2vox}, point-clouds~\cite{ma2024mfp3d, fan2017point}, and meshes~\cite{kanazawa2018learning, wang2018pixel2mesh} to recover and represent lost 3D information from 2D images at the cost of requiring multi-view inputs or sacrificing details.
Neural Radiance Fields (NeRF)~\cite{mildenhall2020nerf} revolutionized 3D reconstruction by using a network to represent a Signed Distance Field with many implementations~\cite{wang2022nerfneuralradiancefields, Pumarola_2021_CVPR, Deng_2022_CVPR}. More recently, 3D Gaussian Splatting~\cite{kerbl20233d} produced high quality 3D meshes by using many Gaussians to represent a scene.  
Additionally, diffusion-based models~\cite{Ho2020Diffusion, Rombach_2022_CVPR} synthesize novel views from a single image~\cite{liu2023one-2-3-45, realfusion2023, deng2023nerdi, metzer2023latent}, and these images serve as inputs to the reconstruction pipeline.

Despite these advances, the real-world scale is not captured in these reconstruction methods, limiting their usability in physically grounded applications. NeRF and Gaussian Splatting rely on multiple inputs, lack scale, and require extensive optimization for reconstruction. Diffusion-based approaches work on single-image 3D reconstruction but have an ill-defined nature that cannot capture high-quality geometry with physical scale accuracy. To address these limitations, our proposed method not only produces a single-view 3D reconstruction which captures the geometry of the object from a single image, but also infers real-scale making the framework suitable for real-world applications.

\subsection{Volume Estimation}
Volume estimation is crucial in many domains~\cite{medical_image_processing, liu2005system, Dehais2017TwoViewFoodReconstruction, he2024MetafoodChallenge, e-commerce3d, remondino20143d, robotics_3d, autonomous_3d, manufacturing_3d, Vinod20243DFoodPortion} where 3D or volumetric information is essential. 
Various approaches have been developed for volume estimation, especially in applications like nutritional analysis. These include depth-based methods~\cite{lo2019depth, Shao2023VoxelReconstruction}, model-based methods~\cite{xu2013model, jia20123d}, stereo image methods~\cite{puri2009recognition}, and neural network-based methods~\cite{vinod2022image, Shao2023VoxelReconstruction, thames2021nutrition5k}. Although these methods work well under controlled conditions, they require either multiple views, prior knowledge of object geometry, or specialized depth sensors.

Recent advances in vision-language models (VLMs) have expanded the potential of AI-driven volume estimation. Models like GPT-4o and DeepMind's Flamingo have demonstrated capabilities in visual reasoning, including object attribute estimation~\cite{alayrac2022flamingo, openai2023gpt4}. However, as we demonstrate in our experiments, these models still exhibit high error rates when estimating physical dimensions and volume, as they lack explicit 3D reasoning mechanisms and cannot be accurately used for nutritional analysis of images.

To address these issues, we propose a framework that leverages existing single-view 3D reconstruction methods to produce real-scale 3D models. Our approach not only reconstructs shapes from single images but also accurately estimates their physical volume, outperforming existing volume estimation techniques, and broadening the scope of applications.

%% file: sections/3_method.tex
\section{Method}
\label{sec:method}
Our approach is built on two key ideas. First, modern visual encoders capture rich semantic and geometric cues from an input image, and these cues implicitly correlate with the object’s real-world scale. Second, the discrepancy between the reconstructed model and the true object size can be learned by comparing image features with geometric attributes of the reconstruction through rendered views. As illustrated in Figure~\ref{fig:method_overview}, our framework consists of two main modules: a single-view 3D reconstruction module, and a Real-Scale Module that learns to scale the obtained 3D model.

\subsection{3D Reconstruction Module}\label{subsec:3d_recon}
Our 3D reconstruction module leverages a zero-shot, single-view method to generate a 3D model from a single input image without any fine-tuning. Specifically, we adopt the One-2-3-45 framework~\cite{liu2023one-2-3-45} due to its remarkable performance in reconstructing 3D shapes and apply it to our framework.

We first provide One-2-3-45 with a segmented input produced by the Segment Anything Model (SAM)~\cite{kirillov2023segment}, enabling the system to synthesize multiple novel views and fuse them to produce a coherent 3D representation of the object. While this reconstruction yields high-quality geometry and appearance, its scale remains arbitrary and does not reflect the object’s true physical dimensions. This scale ambiguity is inherent to all single-view reconstruction methods, including One-2-3-45~\cite{liu2023one-2-3-45}, and prevents direct use of the reconstructed output for any application that relies on real-world measurements. 

\subsection{Real-Scale Module}
The Real-Scale Module addresses the inherent scale ambiguity in single-view 3D reconstructions. The core premise of our approach is that meaningful learned features should capture volumetric cues relevant to real-world object size. Therefore, we utilize powerful 2D feature extractors while retaining shape and perspective of the 3D model by extracting features from multiple rendered views of the reconstructed model.

To obtain a complete view of the 3D model through rendered images from varying angles, we load the model into Blender~\cite{Blender} positioned at the origin. We define a spherical coordinate system for camera placement, allowing unrestricted 360$\degree$ motion around the object. The polar angles in the spherical coordinate system describe the camera's elevation as compared to the $XY$-plane.
The camera is positioned at three fixed polar angles, we choose $-45\degree$, $0\degree$, and $45\degree$, which align the camera slightly below the object looking up at it, in-line with the object, and slightly above the object looking down on it, respectively.
For each polar angle, the camera is rotated around the object at evenly spaced azimuth angles for a full $360\degree$ rotation. This process produces rendered images for each model that capture the object from diverse perspectives while maintaining the 2D structure required for feature extraction. We choose 75 total rendered images per 3D model for training, and we show experimentally that the number of rendered images plays a role in influencing the accuracy of our method.

\begin{figure}[h!]
    \centering
    \includegraphics[width=\linewidth]{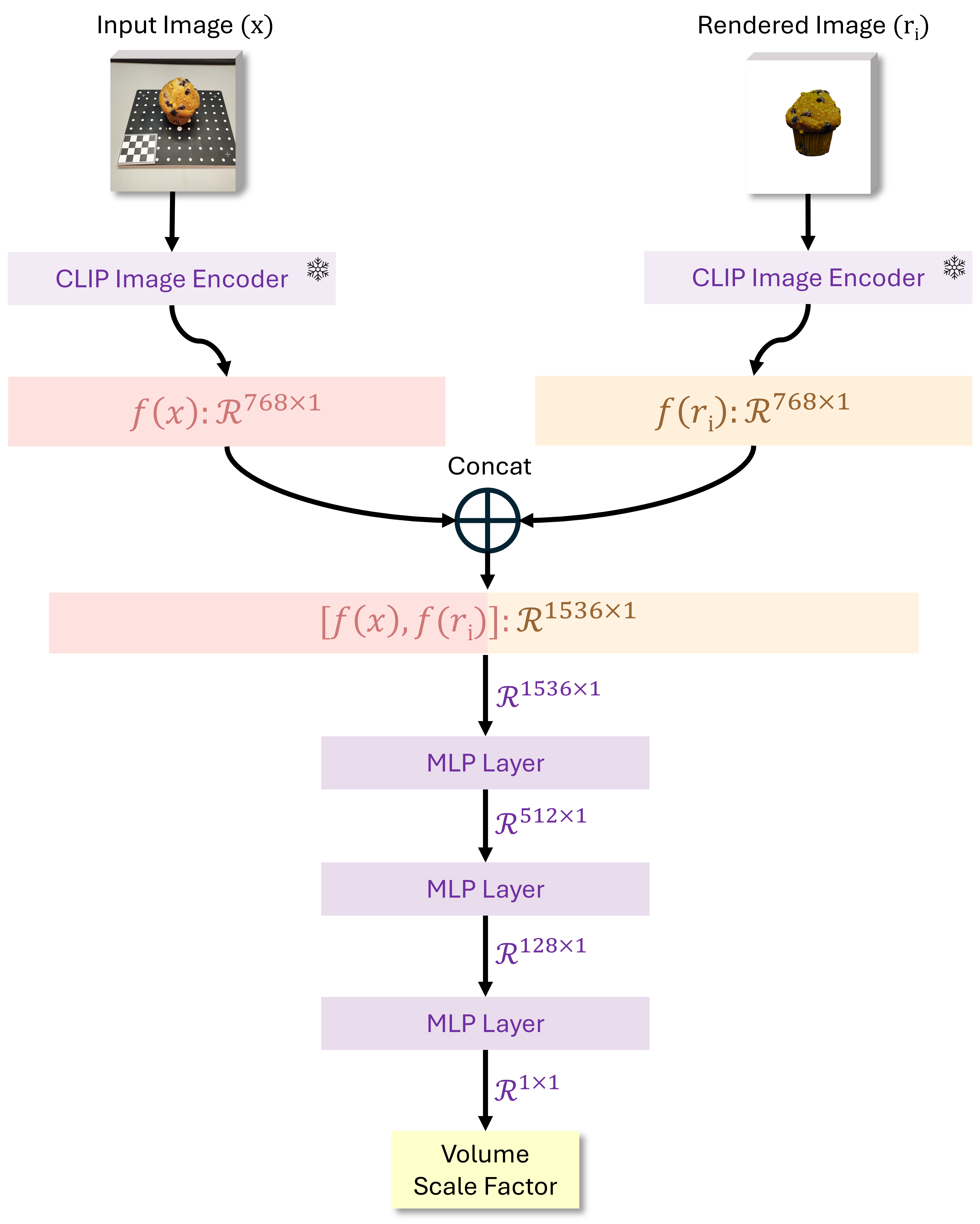}
    \caption{\textbf{Real-Scale Module.} The features from the input image $x$ and the rendered image $r_i$ are extracted and then concatenated to a $1536\times1$ feature vector. The 3 MLP layers bring down the dimensionality and finally output a single regression value which is the volume scale factor.}
    \label{fig:real_scale_module}
\end{figure}

Next, we use these rendered images and the input image for feature extraction via the CLIP model~\cite{radford2021learning}. Although CLIP is not explicitly trained to represent metric scale, its embeddings capture information correlated with an object’s typical size, material, proportions, and functional category. When combined with the geometric cues from the rendered views, these features provide a meaningful signal for scale estimation. Both the input image $x$ and each rendered image $r_i$ (for $i=1,\dots,75$) are passed through a frozen CLIP~\cite{radford2021learning} image encoder $f(\cdot)$ with a ViT-L/14~\cite{dosovitskiy2021vit} backbone to obtain rich feature embeddings $f(x)$ and $f(r_i)$. 

The image embeddings are then concatenated to form an embedding pair as $[f(x), f(r_i)]$ which is then used as inputs to MLP layers that output a singular volume scale factor. The target scale factor is computed by using the ground-truth volume of the object $V_{\textit{gt}}$ and the volume of the reconstructed model $V_{\textit{reconstructed}}$, to obtain:
\begin{equation}\label{eq:1}
    v_{\textit{scale}} = \frac{V_{\textit{gt}}}{V_{\textit{reconstructed}}}
\end{equation} 

Figure~\ref{fig:real_scale_module} shows the architecture of the real-scale module for image feature extraction, feature adaptation, and scale factor regression. Next, each of the $m=75$ rendered images are paired with the input image and yields an individual scale factor $\hat{v}_{\textit{scale}}^i$ $\forall i \in \{1, \dots, 75\}$ after being passed through the MLP layers. The final estimated scale factor is the average scale factor over all the views for each 3D model:
\begin{equation}
    \hat{v}_{\textit{scale}} = \frac{1}{m} \sum_{i=1}^m \hat{v}_{\textit{scale}}^i \: \: \: \: \forall i \in \{1,\dots,m\}
\end{equation}

The model is trained to estimate the target volume scale factor $v_{\textit{scale}}$ in Equation~\ref{eq:1}. To address small values in the ground truth volume that lead to a large percentage error (\textit{e.g.}, an estimate of 4 mL to a target of 1 mL is a 400\% error), we use a normalized version standard L1 loss function:
\begin{equation} \label{eq:4}
    \mathcal{L}_{\textit{regression}} = \frac{1}{mN}\sum_{i=1}^N \sum_{j=1}^m \frac{1}{v_{\textit{scale}}} |{v_{\textit{scale}}} - \hat{v}_{\textit{scale}}^j|
\end{equation} 
where $N$ is the number of images and 3D models in the dataset and $\hat{v}_{\textit{scale}}^j$ is the estimated volume scale factor for the $j$-th input image and render pair. 
This weighting penalizes errors more heavily when the target value is small, which improves model stability and accuracy.

Finally, the scaling is done by using the reconstructed mesh from the 3D Reconstruction Module (Section~\ref{subsec:3d_recon}) $M_\text{recon}$ and applying the estimated scale factor $\sqrt[3]
{\hat{v}_\text{scale}}$ as a scale transformation matrix to bring the model to real-world physical units. We evaluate our approach by comparing its volume estimation performance against existing methods, thereby demonstrating the effectiveness of our framework in achieving real-scale reconstruction.

%% file: sections/4_experimental_results.tex
\section{Experimental Results}
\label{sec:experimental_results}
\subsection{Volume Estimation}
Volume estimation provides a quantifiable metric to evaluate the accuracy of our real-scale 3D reconstructions, which is crucial for many applications. We demonstrate the efficacy of our method on two publicly available datasets, OmniObject3D~\cite{wu2023omniobject3d} to show that our method can work on volume estimation of general objects, and MetaFood3D~\cite{chen2024metafood3d} to show its performance for nutritional analysis against several baselines, including learning-based approaches, 3D-assisted methods, and vision-language models (VLMs), showing that our method successfully reconstructs real-scale 3D models from single-view images.

\begin{table*}[ht!]
    \centering
    \renewcommand{\arraystretch}{1.1}
    \setlength{\tabcolsep}{5pt}
    \resizebox{\textwidth}{!}{
    \begin{tabular}{r | c c c c c c c c}
    \toprule
        \multirow{2}{*}{\textbf{Metrics}} & \multicolumn{7}{c}{\textbf{Volume Estimation Methods}} \\
        & Baseline  & Category Mean & RGB Est.~\cite{thames2021nutrition5k}  & Depth Recon.~\cite{fang2016comparison}  & 3D Assisted~\cite{Vinod20243DFoodPortion}   & GPT-4o~\cite{achiam2023gpt}    & GPT-4o (w/context)~\cite{wei2022chain}   & \textbf{Ours} \\
        \midrule
        & \multicolumn{7}{c}{\textbf{MetaFood3D Dataset}} \\
        MAE (mL)    $\downarrow$    & 151.85    & 134.86 & 161.70    & 199.60        & 123.34        & 84.95     & \underline{83.65}                 & \textbf{59.09} 
 (\textbf{-29.3\%}) \\
        MAPE (\%)   $\downarrow$    & 845.69  & 202.88  & 86.84     & 277.65        & 104.07         & 153.80    & \underline{63.25}                 & \textbf{35.83} (\textbf{-43.3\%}) \\
        r           $\uparrow$      & 0   & 0.46      & \underline{0.76}      & 0.68          &  0.42         & 0.58      &  0.67                 & \textbf{0.87} (\textbf{+14.5\%}) \\
        $R^2$     $\uparrow$    &      0    & 0.07 &\underline{0.58} & 0.47 & 0.17 &0.34 & 0.45 & \textbf{0.75} (\textbf{+29.3\%}) \\
        $\cos({\theta})$ $\uparrow$ & 0.707  & 0.63 & 0.79 & \underline{\textbf{0.99}} & 0.65 & 0.94 & 0.95 & \textbf{0.99} (\textbf{+0\%}) \\
        \midrule
        & \multicolumn{7}{c}{\textbf{OmniObject Dataset}} \\
        MAE (mL) $\downarrow$  & 179.49 & 118.47 & \underline{91.38}  & 225.42 & - & 136.69 & 112.61  & \textbf{70.49} (\textbf{-22.86\%}) \\
        MAPE (\%) $\downarrow$ & 844.29 & \underline{87.46} & 158.74  & 87.84 & -  & 169.81 & 155.28 & \textbf{85.96} (\textbf{-2.14\%}) \\
        r $\uparrow$ &   0   & 0.63 & \underline{0.71}  &   0.32    &  -      &   0.40  &    0.52    &  \textbf{0.94} (\textbf{+32.39\%}) \\
        $R^2$     $\uparrow$    &      0  & 0.35   & 0.38 & \underline{0.53} & - & 0.13 & 0.40 & \textbf{0.88} (\textbf{+66.03\%}) \\
        $\cos({\theta})$ $\uparrow$ & 0.707  & 0.76 & 0.76 & 0.74 & - & 0.63 & \underline{0.76} & \textbf{0.94} \textbf{(+22.07\%)} \\
        \bottomrule
    \end{tabular}}
    \caption{\textbf{Volume Estimation Comparison.} We evaluate performance on the MetaFood3D and OmniObject3D datasets. Our method consistently outperforms all baselines, achieving up to a \textbf{29.3\% reduction in MAE} and a \textbf{43.3\% reduction in MAPE} compared to the next-best methods (underlined). The “Baseline” refers to predicting the dataset's mean volume for all inputs. The ``Category Mean'' predicts the mean volume of the category for each item belonging to that category. ``3D Assisted'' is omitted from OmniObject3D as it requires a checkerboard pattern absent in that dataset. } 
    \vspace{-0.5cm}
    \label{tab:volume_comparison}
\end{table*}
\subsubsection{Dataset}
We perform experiments on two datasets: 
\begin{itemize} 
\item \textbf{MetaFood3D}~\cite{chen2024metafood3d}: Contains 637 3D food items with real physical dimensions and nutritional annotations. We use 535 3D models for training and 102 for testing. 
\item \textbf{OmniObject3D}~\cite{wu2023omniobject3d}: A total of 3,417 3D models of generic objects are used, with 2,763 in the training set and 654 in the testing set.
\end{itemize} 
The train-test split is stratified so that each category (\textit{e.g.}, Apples) has samples in both the training and testing set. 

For the single-view 3D reconstruction, one frame from an RGB video is randomly sampled during inference, while 10 frames are sampled per item during training to enrich the model’s perspective. With 75 rendered views per frame, this results in a large number of input-render pairs (\textit{e.g.}, 396,750 pairs for MetaFood3D and 2,072,250 training pairs for OmniObject3D).

\subsubsection{Metrics}
To evaluate our model's performance, we rely on standard regression metrics - Mean Absolute Error (MAE), Mean Absolute Percentage Error (MAPE), Pearson Coefficient ($r$), Coefficient of Determination $R^2$, and cosine similarity $\cos(\theta)$ which are defined as:

\vspace{-0.4cm}
{\small
\begin{alignat*}{2}
    &\text{MAE} = \frac{1}{N}\sum_{i=1}^N |\hat{\mathbf{V}}_{\textit{est}}^i - \mathbf{V}_{\textit{gt}}^i|, \quad
    &\text{MAPE} = \frac{1}{N}\sum_{i=1}^N \frac{|\hat{\mathbf{V}}_{\textit{est}}^i - \mathbf{V}_{\textit{gt}}^i|}{\mathbf{V}_{\textit{gt}}^i}
\end{alignat*}
}
\vspace{-0.5cm}

{\small
\begin{align*} 
&r = \frac{\sum_i (\hat{\mathbf{V}}_{\textit{est}}^i - \bar{\mathbf{V}}_{\textit{est}})(\mathbf{V}_{\textit{gt}}^i - \bar{\mathbf{V}}_{\textit{gt}})} {\sqrt{\sum_i{(\hat{\mathbf{V}}_{\textit{est}}^i - \bar{\mathbf{V}}_{\textit{est}})}^2{\sum_i {(\mathbf{V}_{\textit{gt}}^i - \bar{\mathbf{V}}_{\textit{gt}})}^2}}}
\end{align*}
\begin{alignat*}{2}
    & R^{2} &= 1 - \frac{\sum_i \big(\mathbf{V}_{\text{gt}}^i - \hat{\mathbf{V}}_{\text{est}}^i)^2}{\sum_i \big(\mathbf{V}_{\text{gt}}^i - \bar{\mathbf{V}}_{\text{gt}})^2}, \quad
    &\cos(\theta) = \frac{\mathbf{V}_{\text{est}} \cdot \mathbf{V}_{\text{gt}}}{\norm{\mathbf{V}_{\text{est}}} \norm{\mathbf{V}_{\text{gt}}}}
\end{alignat*}
}
where $N$ is the number of 3D models in the dataset, $\hat{\mathbf{V}}_{\textit{est}}$ are our method's volume estimates, $\mathbf{V}_{\textit{gt}}$ are the ground truth volumes of the objects, $\bar{\mathbf{V}}_{\textit{est}}$ is the mean of the estimated volumes across the dataset, and $\bar{\mathbf{V}}_{\textit{gt}}$ is the mean of the ground-truth 3D volumes across the dataset.

The units of volume in both datasets are scaled to milliliters (mL). The MAPE is used to determine the best-performing method mainly due to the prevalence of low ground-truth values in the datasets, which causes spikes in the MAPE. The Pearson coefficient ($r$) and the coefficient of determination ($R^2$) assess the correlation between the estimates and ground truth, and cosine similarity measures the alignment of the estimation trend with the ideal $y=x$ line.

\subsubsection{Implementation Details}
An advantage of our method is that the weights of the CLIP feature encoder~\cite{radford2021learning} are frozen, therefore, we can separate the feature extraction from the MLP training, thus considerably reducing the training time and resources since each image does not have to be passed through large transformer models for each iteration during training. 

The MLP layers are supervised on the modified regression loss (Equation~\ref{eq:4}) and run for 300 epochs. The batch size is set to 64 with a learning rate of $10^{-4}$. A learning rate scheduler is used to decrease the learning rate by a factor of 0.7 every 10 epochs. Training and inference are run on 4 NVIDIA 1080 Ti GPUs.

\subsubsection{Methods of Comparison}
The following methods that have been shown to have strong performance in volume estimation tasks are chosen to compare with our proposed method:

\textbf{RGB Estimation}~\cite{thames2021nutrition5k} - A baseline model that directly regresses volume from a single RGB image using a lightweight feature extractor~\cite{he2016deep} followed by MLP layers.

\textbf{Depth Reconstruction}~\cite{fang2016comparison} - Generates a depth map from the RGB image using a monocular depth estimator, Metric3D~\cite{yin2023metric3d, hu2024metric3d}, then reconstructs a voxel model. Ground-truth volumes are used to calibrate the output scale to the dataset's scale via a category wise mean scale factor.

\textbf{3D Assisted Estimation}~\cite{Vinod20243DFoodPortion} - Matches the input image to a known 3D template model using geometric cues and a reference checkerboard for scale. Not applicable to OmniObject3D~\cite{wu2023omniobject3d} due to the lack of reference markers.

\textbf{GPT-4o} - Uses the GPT-4o multimodal model~\cite{openai_gpt_api} to estimate volume from a single image. The prompt instructs the model to output only the object’s volume (in mL), relying purely on visual input.

\textbf{GPT-4o (w/ context)} - Extends GPT-4o by including textual context in the prompt (\textit{e.g.}, “an image of a banana”) to improve estimation accuracy~\cite{brown2020language, wei2022chain}, leveraging prior knowledge of the object class.

\begin{figure}[h]
    \centering
    \includegraphics[width=\linewidth]{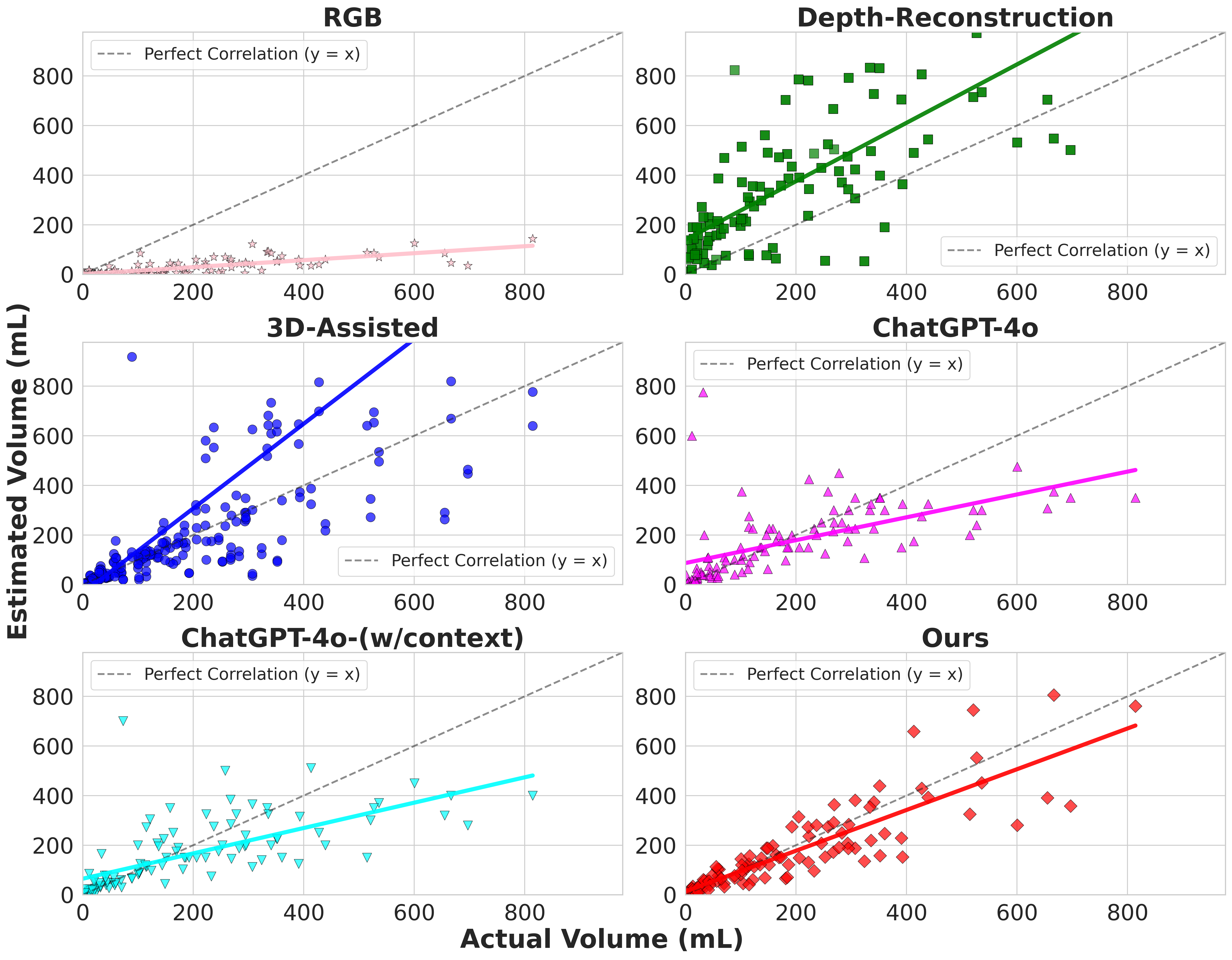}
    \caption{\textbf{Estimated Volume vs Actual Volume.} The predictions of each volume estimation method are compared with the ground-truth volumes, with perfect scores falling on the $y=x$ line. A linear fit line for each method shows the distance of the predictions from the $y=x$ line. The linear fit lines illustrate the deviation of predictions from the ground truth. Our method exhibits the best fit, indicating superior accuracy.}
    \label{fig:gt_v_est}
\end{figure}

Table~\ref{tab:volume_comparison} presents a quantitative comparison of our method against multiple baselines, including traditional volume estimation approaches and vision-language models (GPT-4o with and without additional context). 
All the trainable methods have been trained or fine-tuned on the respective datasets that it was tested on. The reconstruction-based methods used the categorical mean ground-truth volume scaling factor to bring them to physical units.
Our method achieves the lowest absolute and percentage errors and highest correlation metrics across both datasets, improving the absolute error by nearly 30\%. These results underscore the critical role of real-scale 3D reconstruction in accurate volume estimation. 
To verify that our model learns generalizable scale priors rather than overfitting to specific categories, we conducted additional experiments despite limited data availability. In a random, non-stratified split of MetaFood3D~\cite{chen2024metafood3d}, our method demonstrated strong generalization with 82.46\% MAPE, improving nearly 4-fold over the RGB Only method (299.52\% MAPE). Furthermore, in a cross-dataset evaluation (training on OmniObject3D~\cite{wu2023omniobject3d} and testing on MetaFood3D~\cite{chen2024metafood3d}), our method yielded a significantly lower MAE (104.08 mL) than the RGB Only (174.57 mL), indicating effective transfer of learned scale features.

\subsection{Performance Analysis}
\begin{figure}[h]
    \centering
    \includegraphics[width=\linewidth]{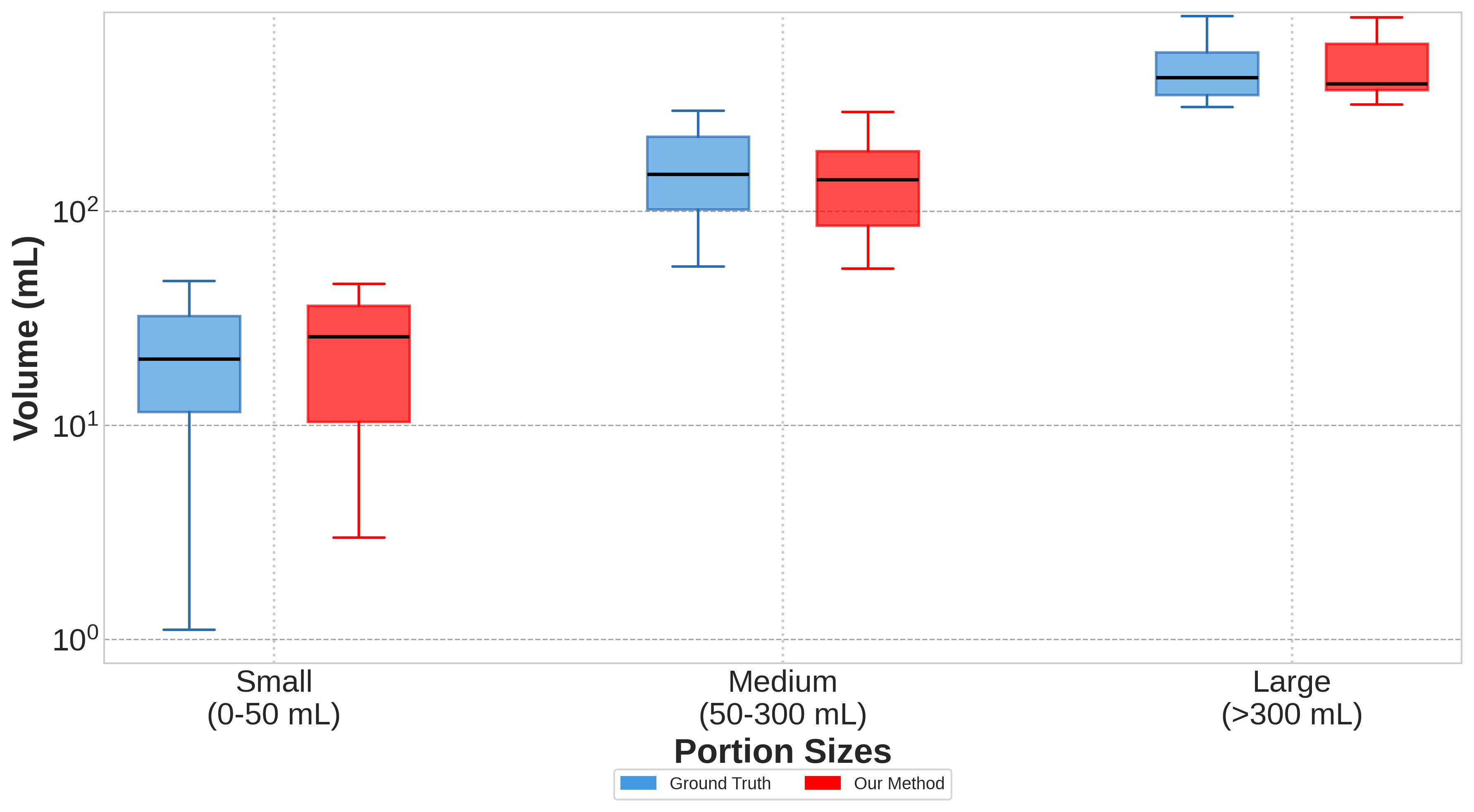}
    \caption{\textbf{Prediction distribution across portion sizes.} Our method's estimated volumes are compared with ground-truth values for small (0 - 50 mL), medium (50 - 300 mL), and large ($>$300 mL) portions. The close match between the distributions demonstrates the accurate real-scale predictions of our approach across diverse object sizes.}
    \label{fig:gt_v_ours_boxplot}
\end{figure}

Our model outperforms existing methods across all metrics and datasets. To better understand performance, we group ground-truth volumes in the MetaFood3D dataset into three categories: small ($0-50$ mL), medium ($50-300$ mL), and large ($>300$ mL). As shown in Figure~\ref{fig:gt_v_ours_boxplot}, our prediction distribution closely aligns with the ground truth across all volume distributions, highlighting the model’s ability to generalize across volume scales. Table~\ref{tab:energy_estimation} demonstrates a real-world use case on the MetaFood3D~\cite{chen2024metafood3d} dataset, where our estimated volumes are converted to food energy (kCal) using USDA's food codes~\cite{montville2013usda}. Despite relying on only a single image, our method significantly outperforms others in energy estimation accuracy.

\begin{table}[h]
    \centering
    \begin{adjustbox}{max width=\linewidth}
    \begin{tabular}{l|cc|cc}
        \toprule
        \multirow{2}{*}\textbf{Method} & \multicolumn{2}{c}{\textbf{Volume (mL)}} & \multicolumn{2}{c}{\textbf{Energy (kCal)}} \\
         & \textbf{MAE} & \textbf{MAPE} (\%) &\textbf{MAE} & \textbf{MAPE} (\%)\\
        \midrule
        Baseline & 165.75 & 836.50  & 214.55 & 1135.93 \\
        Stereo Recon. & 153.58 & 214.95 & 262.07 & 244.80  \\
        Voxel Recon. & 120.16 & 96.31  & 174.45 & 130.16 \\
        3D Assisted & 186.45 & 83.26 & 287.11 &  132.42 \\
        Ours & \textbf{61.24} & \textbf{37.82}  & \textbf{163.7} &  \textbf{42.73} \\
        \bottomrule
    \end{tabular}
    \end{adjustbox}
    \caption{\textbf{Comparison of image-based dietary assessment methods with our method.} Our method outperforms existing methods (Stereo Recon.~\cite{Dehais2017TwoViewFoodReconstruction} Voxel Recon.~\cite{fang2016comparison}, 3D Assisted~\cite{Vinod20243DFoodPortion}) in volume and energy estimation using while only a single image input. The MAE and MAPE are significantly lower than other methods.}
    \label{tab:energy_estimation}
\end{table}

\subsection{Ablation Studies}
To validate the design choices of our framework, we conduct ablation studies to assess the contributions of individual components: pairwise image feature extraction, image encoder selection,  number of render images incorporated, and 3D reconstruction methods.

\subsubsection{Pairwise Image Feature Extraction}
Our method combines the features of the input image and the rendered image for accurate volume estimation. This joint representation enables the model to learn the scale factor by comparing the original image with the reconstructed shape. To validate this, we test three configurations in the same experimental setup: using only input image features ($f(x)$), only rendered image features ($f(r)$), and combined image features ($[f(x), f(r)]$).

As shown in Table~\ref{tab:pairwise_results_table}, our method's unique concatenation of features yields significantly lower error (MAE: 59.09 mL, MAPE: 35.83\%) compared to using either feature alone.

\begin{table}[h]
    \centering
    \small
    \begin{tabular}{l|c c}
    \toprule
       \textbf{Input}  & \textbf{MAE (mL)} &\textbf{ MAPE (\%)} \\
       \midrule
        $f(x)$  & 124.94 & 76.10\\
        $f(r)$ & 153.60 & 90.80\\
        \textbf{$[f(x), f(r)]$} & \textbf{59.09} & \textbf{35.83}\\
        \bottomrule
    \end{tabular}
    \caption{\textbf{Effect of feature pairing.} Combining input image features $f(x)$ with rendered image features $f(r)$ significantly improves volume estimation accuracy over using either component alone.}
    \label{tab:pairwise_results_table}
\end{table}

\subsubsection{Image Encoder Analysis}
The efficacy of our pipeline relies on the features extracted through the image encoder of the CLIP model~\cite{radford2021learning}. We evaluate the contribution of this feature extractor to our pipeline by using different backbone networks, both CLIP and non-CLIP, and their variants. 

\begin{table}[h]
    \centering
    \small
    \begin{tabular}{l|c c}
    \toprule
       \textbf{Backbone}  & \textbf{MAE (mL)} &\textbf{ MAPE (\%)} \\
       \midrule
        DeiT Small & 127.47 & 67.75 \\
        DeiT Base & 90.40 & 52.77 \\
        ViT B/32 & 95.54 & 59.56 \\
        ViT L/14 & 91.06 & 48.59 \\
        ResNet50 - CLIP & 77.08 & 50.72 \\
        ResNet101 - CLIP & 76.57 & 55.20 \\
        ViT B/32 - CLIP & 93.44 & 49.11 \\
        ViT L/14 - CLIP & 74.33 & 50.78 \\
        \textbf{ViT L/14@336px - CLIP} & \textbf{59.09} & \textbf{35.83} \\
        \bottomrule
    \end{tabular}
    \caption{\textbf{Comparison of feature extractors.} Transformer-based models, especially the CLIP ViT-L/14@336px variant, provide the most discriminative features for volume estimation.}
    \label{tab:backbone_analysis}
\end{table}

Table~\ref{tab:backbone_analysis} shows the CLIP~\cite{radford2021learning} with the ViT-L/14~\cite{dosovitskiy2021vit} variant outperforms all other models. The CLIP-trained models mostly outperform its original counterparts.
However, the range of MAE and MAPE for most models do not vary significantly showing the capability of these strong feature extractors in volume estimation.

\subsubsection{Impact of the Number of Rendered Images}
To understand the impact of the number of rendered images used for our method, we evaluate $m \in {5,15,25,50,75}$, which is the number of rendered images used during inference on MetaFood3D (Table~\ref{tab:k_renders}). MAPE improves steadily up to $m=25$, achieving its minimum of $35.83 \%$, allowing us to use fewer images ($m=25$) to save on computation during inference while maintaining high accuracy.

\begin{table}[h]
    \centering
    \small
    \begin{tabular}{l|c c}
    \toprule
       \textbf{\textbf{m}-Renders}  & \textbf{MAE (mL)} &\textbf{ MAPE (\%)} \\
       \midrule
        5 & 70.76 & 48.34 \\
        15 & 63.51 & 43.49 \\
        \textbf{25} & \textbf{59.09} & \textbf{35.83} \\
        50 & 60.88 & 40.48 \\
        75 & 59.52 & 38.42 \\
        \bottomrule
    \end{tabular}
    \caption{\textbf{Effect of Number of Rendered Views ($m$) on Volume Estimation.} We evaluate how varying the number of rendered views used during inference affects performance. Accuracy improves up to $m=25$, after which additional views lead to insignificant or even decrease in performance.}
    \label{tab:k_renders}
\end{table}

\subsubsection{3D Reconstruction Method}
We show that our framework is flexible to the 3D reconstruction method by testing our framework on other reconstruction methods. 

\begin{table}[h]
    \centering
    \small
    \begin{adjustbox}{max width=\linewidth}
    \begin{tabular}{l|c c}
    \toprule
       \textbf{3D Reconstruction}  & \textbf{MAE (mL)} &\textbf{ MAPE (\%)} \\
       \midrule
        One-2-3-45~\cite{liu2023one-2-3-45} & 59.09 & 35.83 \\
        TripoSR~\cite{tochilkin2024triposr} & 88.80 & 54.00 \\
        \bottomrule
    \end{tabular}
    \end{adjustbox}
    \caption{\textbf{Impact of 3D reconstruction methods.} Our framework demonstrates flexibility by delivering accurate volume estimation with different single-view 3D reconstruction methods, with One-2-3-45 yielding superior performance.}
    \label{tab:reconstruction_analysis}
\end{table}

Table~\ref{tab:reconstruction_analysis} highlights the importance of the 3D reconstruction method in volume estimation. Our framework achieves better volume estimation when integrated with One-2-3-45 (MAE: 59.09 mL, MAPE: 34.83\%) compared to TripoSR (MAE: 88.80 mL, MAPE: 54.00\%). This suggests that One-2-3-45 has higher-quality reconstructions for this dataset that lead to more accurate scale estimations, and importantly, that our method is flexible enough to work with different reconstruction techniques and still achieve accurate volume estimation.

%% file: sections/5_conclusion.tex
\section{Discussion and Conclusion}
\label{sec:conclusion}
In this work, we introduced a Real-Scale 3D Reconstruction framework that leverages pairwise image feature extraction for accurate volume estimation, enabling the scaling of reconstructed 3D models to their true physical dimensions. Our experiments show that both the quality of the feature extraction backbone and the integration of features from the input image and its rendered views are critical to capturing the object's size information accurately. By integrating semantic priors from CLIP with geometric cues from rendered views, our method bridges the gap between generative 3D vision and real-world applicability. Experimental results on MetaFood3D and OmniObject3D demonstrate a 30\% reduction in volume estimation error compared to existing baselines.

Our method distinguishes itself from traditional 3D reconstruction techniques, which often neglect physical scale, and from vision-language models (VLMs) that, while promising, lack precise spatial reasoning. By learning a scale factor directly from rich visual features, our approach bridges this gap, achieving a significant reduction in error compared to existing volume estimation methods. We also show the utility of our method on nutritional analaysis, further enhancing the field of precision nutrition. Beyond precision nutrition, where we achieve high accuracy in caloric estimation, our approach offers a unique solution for scale-aware 3D modeling in broader domains such as e-commerce and robotics. Future work will focus on integrating this framework into mobile platforms for real-time dietary monitoring.

%% file: references.bib
@String(CVPR= {IEEE Conf. Comput. Vis. Pattern Recog.})

@String(ECCV= {Eur. Conf. Comput. Vis.})

@String(ICME = {Int. Conf. Multimedia and Expo})

@String(ICIP = {IEEE Int. Conf. Image Process.})

@String(CVPRW= {IEEE Conf. Comput. Vis. Pattern Recog. Worksh.})

@String(CVPR  = {CVPR})

@String(ECCV  = {ECCV})

@String(ICME  =	{ICME})

@String(ICIP  = {ICIP})

@String(CVPRW= {CVPRW})

@article{liu2023one-2-3-45,
  title = {One-2-3-45: Any Single Image To 3D Mesh In 45 Seconds Without Per-Shape Optimization},
  author = {Liu, Minghua and Xu, Chao and Jin, Haian and Chen, Linghao and Varma T, Mukund and Xu, Zexiang and Su, Hao},
  journal = {Proceedings Of The 2023 Advances In Neural Information Processing Systems},
  volume = {36},
  pages = {22226--22246},
  year = {2023}
}

@article{liu2005system,
  title = {A System For Brain Tumor Volume Estimation Via MR Imaging And Fuzzy Connectedness},
  author = {Liu, Jianguo and Udupa, Jayaram K and Odhner, Dewey and Hackney, David and Moonis, Gul},
  journal = {Computerized Medical Imaging And Graphics},
  volume = {29},
  number = {1},
  pages = {21--34},
  year = {2005}
}

@article{medical_image_processing,
  title = {Research On The Application Of Deep Learning In Medical Image Segmentation And 3D Reconstruction},
  author = {Zi, Yun and Wang, Qi and Gao, Zijun and Cheng, Xiaohan and Mei, Taiyuan},
  journal = {Academic Journal Of Science And Technology},
  volume = {10},
  number = {2},
  pages = {8--12},
  year = {2024}
}

@ARTICLE{Dehais2017TwoViewFoodReconstruction,
  author = {Dehais, Joachim and Anthimopoulos, Marios and Shevchik, Sergey and Mougiakakou, Stavroula},
  journal = {IEEE Transactions On Multimedia},
  title = {Two-View 3D Reconstruction For Food Volume Estimation},
  year = {2017},
  volume = {19},
  number = {5},
  pages = {1090-1099},
  doi = {10.1109/TMM.2016.2642792}
}

@article{he2024MetafoodChallenge,
  title = {MetaFood CVPR 2024 Challenge On Physically Informed 3D Food Reconstruction: Methods And Results},
  author = {He, Jiangpeng and Chen, Yuhao and Vinod, Gautham and Mahmud, Talha Ibn and Zhu, Fengqing and Delp, Edward and Wong, Alexander and Xi, Pengcheng and AlMughrabi, Ahmad and Haroon, Umair and others},
  journal = {ArXiv Preprint ArXiv:2407.09285},
  year = {2024}
}

@ARTICLE{e-commerce3d,
  author = {Tuan, Thai Thanh and Minar, Matiur Rahman and Ahn, Heejune and Wainwright, John},
  journal = {IEEE Access}, 
  title = {Multiple Pose Virtual Try-On Based On 3D Clothing Reconstruction}, 
  year = {2021},
  volume = {9},
  pages = {114367-114380},
  doi = {10.1109/ACCESS.2021.3104274}
}

@Article{kerbl20233d,
      author       = {Kerbl, Bernhard and Kopanas, Georgios and Leimk{\"u}hler, Thomas and Drettakis, George},
      title        = {3D Gaussian Splatting for Real-Time Radiance Field Rendering},
      journal      = {ACM Transactions on Graphics},
      number       = {4},
      volume       = {42},
      month        = {July},
      year         = {2023},
      url          = {https://repo-sam.inria.fr/fungraph/3d-gaussian-splatting/}
}

@article{tochilkin2024triposr,
  title = {Triposr: Fast 3D Object Reconstruction From A Single Image},
  author = {Tochilkin, Dmitry and Pankratz, David and Liu, Zexiang and Huang, Zixuan and Letts, Adam and Li, Yangguang and Liang, Ding and Laforte, Christian and Jampani, Varun and Cao, Yan-Pei},
  journal = {ArXiv Preprint ArXiv:2403.02151},
  year = {2024}
}

@article{mildenhall2020nerf,
  title = {NeRF: Representing Scenes As Neural Radiance Fields For View Synthesis},
  author = {Mildenhall, Ben and Srinivasan, Pratul P and others},
  journal = {Proceedings Of The 2020 European Conference On Computer Vision},
  year = {2020},
  pages = {405-421}
}

@article{realfusion2023,
  title = {RealFusion: 3D Reconstruction Via Realistic Image Synthesis},
  author = {Johnson, Peter and Wong, Emily},
  journal = {Proceedings Of The 2023 Neural Information Processing Systems},
  year = {2023}
}

@article{radford2021learning,
  title = {Learning Transferable Visual Models From Natural Language Supervision},
  author = {Radford, Alec and Kim, Jong Wook and others},
  journal = {Proceedings Of The 2021 International Conference On Machine Learning},
  year = {2021}
}

@book{remondino20143d,
  title = {3D Recording And Modelling In Archaeology And Cultural Heritage},
  author = {Remondino, Fabio and Campana, Stefano},
  year = {2014},
  publisher = {British Archaeological Reports Oxford}
}

@book{robotics_3d,
  title = {3D Robotic Mapping: The Simultaneous Localization And Mapping Problem With Six Degrees Of Freedom},
  author = {N{\"u}chter, Andreas},
  volume = {52},
  year = {2008},
  publisher = {Springer}
}

@article{autonomous_3d,
  title = {Are We Ready For Autonomous Driving? The Kitti Vision Benchmark Suite},
  author = {Geiger, Andreas and Lenz, Philip and Urtasun, Raquel},
  journal = {Proceedings Of The 2012 IEEE Conference On Computer Vision And Pattern Recognition},
  pages = {3354--3361},
  year = {2012},
  organization = {IEEE}
}

@article{manufacturing_3d,
  title = {Applications Of 3D Scanning And Reverse Engineering Techniques For Quality Control Of Quick Response Products},
  author = {Yao, AWL},
  journal = {The International Journal Of Advanced Manufacturing Technology},
  volume = {26},
  pages = {1284--1288},
  year = {2005},
  publisher = {Springer}
}

@article{wu2023omniobject3d,
  title = {Omniobject3d: Large-Vocabulary 3D Object Dataset For Realistic Perception, Reconstruction And Generation},
  author = {Wu, Tong and Zhang, Jiarui and Fu, Xiao and Wang, Yuxin and Ren, Jiawei and Pan, Liang and Wu, Wayne and Yang, Lei and Wang, Jiaqi and Qian, Chen and Others},
  journal = {Proceedings Of The IEEE/CVF Conference On Computer Vision And Pattern Recognition},
  pages = {803--814},
  year = {2023}
}

@article{chen2024metafood3d,
  title = {Metafood3D: Large 3D Food Object Dataset With Nutrition Values},
  author = {Chen, Yuhao and He, Jiangpeng and Czarnecki, Chris and Vinod, Gautham and Mahmud, Talha Ibn and Raghavan, Siddeshwar and Ma, Jinge and Mao, Dayou and Nair, Saeejith and Xi, Pengcheng and Others},
  journal = {ArXiv Preprint ArXiv:2409.01966},
  year = {2024}
}

@article{Shao2023VoxelReconstruction,
  title = {An End-To-End Food Portion Estimation Framework Based On Shape Reconstruction From Monocular Image},
  author = {Shao, Zeman and Vinod, Gautham and He, Jiangpeng and Zhu, Fengqing},
  journal = {Proceedings Of The 2023 IEEE International Conference On Multimedia And Expo (ICME)},
  pages = {942--947},
  year = {2023},
  doi = {10.1109/ICME55011.2023.00166},
  publisher = {IEEE Computer Society},
  address = {Los Alamitos, CA, USA},
  month = {Jul}
}

@article{xie2019pix2vox,
  title = {Pix2Vox: Context-Aware 3D Reconstruction From Single And Multi-View Images},
  author = {Xie, Haozhe and Yao, Hongxun and Sun, Xiaoshuai and Zhou, Shangchen and Zhang, Shengping},
  journal = {Proceedings Of The IEEE/CVF International Conference On Computer Vision},
  pages = {2690--2698},
  year = {2019}
}

@article{ma2024mfp3d,
  title = {MFP3D: Monocular Food Portion Estimation Leveraging 3D Point Clouds},
  author = {Ma, Jinge and Zhang, Xiaoyan and Vinod, Gautham and Raghavan, Siddeshwar and He, Jiangpeng and Zhu, Fengqing},
  journal = {ArXiv Preprint ArXiv:2411.10492},
  year = {2024}
}

@article{fan2017point,
  title = {A Point Set Generation Network For 3D Object Reconstruction From A Single Image},
  author = {Fan, Haoqiang and Su, Hao and Guibas, Leonidas J},
  journal = {Proceedings Of The IEEE Conference On Computer Vision And Pattern Recognition},
  pages = {605--613},
  year = {2017}
}

@article{kanazawa2018learning,
  title = {Learning Category-Specific Mesh Reconstruction From Image Collections},
  author = {Kanazawa, Angjoo and Tulsiani, Shubham and Efros, Alexei A and Malik, Jitendra},
  journal = {Proceedings Of The European Conference On Computer Vision (ECCV)},
  pages = {371--386},
  year = {2018}
}

@article{wang2018pixel2mesh,
  title = {Pixel2Mesh: Generating 3D Mesh Models From Single RGB Images},
  author = {Wang, Nanyang and Zhang, Yinda and Li, Zhuwen and Fu, Yanwei and Liu, Wei and Jiang, Yu-Gang},
  journal = {Proceedings Of The European Conference On Computer Vision (ECCV)},
  pages = {52--67},
  year = {2018}
}

@article{wang2022nerfneuralradiancefields,
  title = {NeRF--: Neural Radiance Fields Without Known Camera Parameters},
  author = {Wang, Zirui and Wu, Shangzhe and Xie, Weidi and Chen, Min and Prisacariu, Victor Adrian},
  journal = {ArXiv Preprint 2102.07064},
  year = {2022}
}

@article{Pumarola_2021_CVPR,
  title = {D-NeRF: Neural Radiance Fields For Dynamic Scenes},
  author = {Pumarola, Albert and Corona, Enric and Pons-Moll, Gerard and Moreno-Noguer, Francesc},
  journal = {Proceedings Of The IEEE/CVF Conference On Computer Vision And Pattern Recognition (CVPR)},
  month = {June},
  year = {2021},
  pages = {10318--10327}
}

@article{Deng_2022_CVPR,
  title = {Depth-Supervised NeRF: Fewer Views And Faster Training For Free},
  author = {Deng, Kangle and Liu, Andrew and Zhu, Jun-Yan and Ramanan, Deva},
  journal = {Proceedings Of The IEEE/CVF Conference On Computer Vision And Pattern Recognition (CVPR)},
  month = {June},
  year = {2022},
  pages = {12882--12891}
}

@article{Ho2020Diffusion,
  title = {Denoising Diffusion Probabilistic Models},
  author = {Ho, Jonathan and Jain, Ajay and Abbeel, Pieter},
  journal = {Proceedings Of The 2020 Advances In Neural Information Processing Systems},
  editor = {Larochelle, Hugo and Ranzato, Marc and Hadsell, Raquel and Balcan, Maria-Fernanda and Lin, Hong},
  pages = {6840--6851},
  publisher = {Curran Associates, Inc.},
  volume = {33},
  year = {2020},
}

@article{Rombach_2022_CVPR,
  title = {High-Resolution Image Synthesis With Latent Diffusion Models},
  author = {Rombach, Robin and Blattmann, Andreas and Lorenz, Dominik and Esser, Patrick and Ommer, Bj{\"o}rn},
  journal = {Proceedings Of The IEEE/CVF Conference On Computer Vision And Pattern Recognition (CVPR)},
  month = {June},
  year = {2022},
  pages = {10684--10695}
}

@article{deng2023nerdi,
  title = {Nerdi: Single-View NeRF Synthesis With Language-Guided Diffusion As General Image Priors},
  author = {Deng, Congyue and Jiang, Chiyu and Qi, Charles R and Yan, Xinchen and Zhou, Yin and Guibas, Leonidas and Anguelov, Dragomir and Others},
  journal = {Proceedings Of The IEEE/CVF Conference On Computer Vision And Pattern Recognition},
  pages = {20637--20647},
  year = {2023}
}

@article{metzer2023latent,
  title = {Latent-NeRF For Shape-Guided Generation Of 3D Shapes And Textures},
  author = {Metzer, Gal and Richardson, Elad and Patashnik, Or and Giryes, Raja and Cohen-Or, Daniel},
  journal = {Proceedings Of The IEEE/CVF Conference On Computer Vision And Pattern Recognition},
  pages = {12663--12673},
  year = {2023}
}

@article{Vinod20243DFoodPortion,
  title = {Food Portion Estimation Via 3D Object Scaling},
  author = {Vinod, Gautham and He, Jiangpeng and Shao, Zeman and Zhu, Fengqing},
  journal = {Proceedings Of The 2024 IEEE/CVF Conference On Computer Vision And Pattern Recognition Workshops (CVPRW)},
  year = {2024},
  pages = {3741--3749},
  doi = {10.1109/CVPRW63382.2024.00378}
}

@article{lo2019depth,
  title = {Depth Estimation Based On A Single Close-Up Image With Volumetric Annotations In The Wild: A Pilot Study},
  author = {Lo, Frank P-W and Sun, Yingnan and Lo, Benny},
  journal = {Proceedings Of The 2019 IEEE/ASME International Conference On Advanced Intelligent Mechatronics (AIM)},
  pages = {513--518},
  year = {2019},
  organization = {IEEE}
}

@article{xu2013model,
  title = {Model-Based Food Volume Estimation Using 3D Pose},
  author = {Xu, Chang and He, Ye and Khanna, Nitin and Boushey, Carol J and Delp, Edward J},
  journal = {Proceedings Of The 2013 IEEE International Conference On Image Processing},
  pages = {2534--2538},
  year = {2013},
  organization = {IEEE}
}

@article{jia20123d,
  title = {3D Localization Of Circular Feature In 2D Image And Application To Food Volume Estimation},
  author = {Jia, Wenyan and Yue, Yaofeng and Fernstrom, John D and Zhang, Zhengnan and Yang, Yongquan and Sun, Mingui},
  journal = {Proceedings Of The 2012 Annual International Conference Of The IEEE Engineering In Medicine And Biology Society},
  pages = {4545--4548},
  year = {2012},
  organization = {IEEE}
}

@article{puri2009recognition,
  title = {Recognition And Volume Estimation Of Food Intake Using A Mobile Device},
  author = {Puri, Manika and Zhu, Zhiwei and Yu, Qian and Divakaran, Ajay and Sawhney, Harpreet},
  journal = {Proceedings Of The 2009 Workshop On Applications Of Computer Vision (WACV)},
  pages = {1--8},
  year = {2009},
  organization = {IEEE}
}

@article{vinod2022image,
  title = {Image Based Food Energy Estimation With Depth Domain Adaptation},
  author = {Vinod, Gautham and Shao, Zeman and Zhu, Fengqing},
  journal = {Proceedings Of The 2022 IEEE 5th International Conference On Multimedia Information Processing And Retrieval (MIPR)},
  pages = {262--267},
  year = {2022},
  organization = {IEEE}
}

@article{kirillov2023segment,
  title = {Segment Anything},
  author = {Kirillov, Alexander and Mintun, Eric and Ravi, Nikhila and Mao, Hanzi and Rolland, Chloe and Gustafson, Laura and Xiao, Tete and Whitehead, Spencer and Berg, Alexander C and Lo, Wan-Yen and Others},
  journal = {Proceedings Of The IEEE/CVF International Conference On Computer Vision},
  pages = {4015--4026},
  year = {2023}
}

@Manual{Blender,
  title = {Blender - A 3D Modelling And Rendering Package},
  author = {{Blender Online Community}},
  organization = {Blender Foundation},
  address = {Blender Institute, Amsterdam},
  year = {2018},
  url = {http://www.blender.org}
}

@article{dosovitskiy2021vit,
  title = {An Image Is Worth 16x16 Words: Transformers For Image Recognition At Scale},
  author = {Dosovitskiy, Alexey and Beyer, Lucas and Kolesnikov, Alexander and Weissenborn, Dirk and Zhai, Xiaohua and Unterthiner, Thomas and Dehghani, Mostafa and Minderer, Matthias and Heigold, Georg and Gelly, Sylvain and Uszkoreit, Jakob and Houlsby, Neil},
  journal = {Proceedings Of The International Conference On Learning Representations},
  year = {2021},
  url = {https://openreview.net/forum?id=YicbFdNTTy}
}

@Manual{openai_gpt_api,
  author = {OpenAI},
  title = {GPT API},
  year = {2023},
  howpublished = {\url{https://platform.openai.com/docs/guides/vision}},
  note = {Accessed: Sep. 13, 2024}
}

@article{brown2020language,
  title = {Language Models Are Few-Shot Learners},
  author = {Brown, Tom and Mann, Benjamin and Ryder, Nick and Subbiah, Melanie and Kaplan, Jared and Dhariwal, Prafulla and Neelakantan, Arvind and Shyam, Pranav and Sastry, Girish and Askell, Amanda and Others},
  journal = {Proceedings Of The 34th International Conference On Neural Information Processing Systems},
  volume = {33},
  pages = {1877--1901},
  year = {2020},
  note = {Article No.: 159}
}

@article{wei2022chain,
  title = {Chain-Of-Thought Prompting Elicits Reasoning In Large Language Models},
  author = {Wei, Jason and Wang, Xuezhi and Schuurmans, Dale and Bosma, Maarten and Chi, Eric and Le, Quoc and Zhou, Denny},
  journal = {Proceedings Of The 36th International Conference On Machine Learning},
  pages = {3759--3774},
  year = {2022},
  organization = {PMLR}
}

@Manual{openai2023gpt4,
  title = {GPT-4 Technical Report},
  author = {OpenAI},
  year = {2023},
  howpublished = {\url{https://openai.com/research/gpt-4}}
}

@article{achiam2023gpt,
  title={Gpt-4 technical report},
  author={Achiam, Josh and Adler, Steven and Agarwal, Sandhini and Ahmad, Lama and Akkaya, Ilge and Aleman, Florencia Leoni and Almeida, Diogo and Altenschmidt, Janko and Altman, Sam and Anadkat, Shyamal and others},
  journal={arXiv preprint arXiv:2303.08774},
  year={2023}
}

@article{alayrac2022flamingo,
  title = {Flamingo: A Visual Language Model For Few-Shot Learning},
  author = {Alayrac, Jean-Baptiste and Donahue, Jeff and Luc, Pauline and Miech, Antoine and Barr, Iain and Hasson, Yana and Lenc, Karel and Mensch, Arthur and Millican, Katherine and Reynolds, Malcolm and Others},
  journal = {Proceedings Of The 2022 Advances In Neural Information Processing Systems},
  volume = {35},
  pages = {23716--23736},
  year = {2022}
}

@article{thames2021nutrition5k,
  title = {Nutrition5K: Towards Automatic Nutritional Understanding Of Generic Food},
  author = {Thames, Quin and Karpur, Arjun and Norris, Wade and Xia, Fangting and Panait, Liviu and Weyand, Tobias and Sim, Jack},
  journal = {Proceedings Of The IEEE/CVF Conference On Computer Vision And Pattern Recognition},
  pages = {8903--8911},
  year = {2021}
}

@article{fang2016comparison,
  title = {A Comparison Of Food Portion Size Estimation Using Geometric Models And Depth Images},
  author = {Fang, Shaobo and Zhu, Fengqing and Jiang, Chufan and Zhang, Song and Boushey, Carol J and Delp, Edward J},
  journal = {Proceedings Of The 2016 IEEE International Conference On Image Processing (ICIP)},
  pages = {26--30},
  year = {2016},
  organization = {IEEE}
}

@article{montville2013usda,
  title = {USDA Food And Nutrient Database For Dietary Studies (FNDDS), 5.0},
  author = {Montville, Janice B and Ahuja, Jaspreet KC and Martin, Carrie L and He, Kaushalya Y and Omolewa-Tomobi, Grace and Steinfeldt, Lois C and Anand, Jaswinder and Adler, Meghan E and LaComb, Randy P and Moshfegh, Alanna},
  journal = {Procedia Food Science},
  volume = {2},
  pages = {99--112},
  year = {2013},
  publisher = {Elsevier}
}

@article{yin2023metric3d,
  title={Metric3D: Towards zero-shot metric 3d prediction from a single image},
  author={Yin, Wei and Zhang, Chi and Chen, Hao and Cai, Zhipeng and Yu, Gang and Wang, Kaixuan and Chen, Xiaozhi and Shen, Chunhua},
  journal={Proceedings of the IEEE/CVF International Conference on Computer Vision},
  pages={9043--9053},
  year={2023},
}

@article{hu2024metric3d,
  title={Metric3d v2: A versatile monocular geometric foundation model for zero-shot metric depth and surface normal estimation},
  author={Hu, Mu and Yin, Wei and Zhang, Chi and Cai, Zhipeng and Long, Xiaoxiao and Chen, Hao and Wang, Kaixuan and Yu, Gang and Shen, Chunhua and Shen, Shaojie},
  journal={IEEE Transactions on Pattern Analysis and Machine Intelligence},
  year={2024},
  publisher={IEEE}
}

@article{wang2023optimal,
  title={Optimal dietary patterns for prevention of chronic disease},
  author={Wang, Peilu and Song, Mingyang and Eliassen, A Heather and Wang, Molin and Fung, Teresa T and Clinton, Steven K and Rimm, Eric B and Hu, Frank B and Willett, Walter C and Tabung, Fred K and others},
  journal={Nature medicine},
  volume={29},
  number={3},
  pages={719--728},
  year={2023},
  publisher={Nature Publishing Group US New York}
}

@article{he2016deep,
  title={Deep residual learning for image recognition},
  author={He, Kaiming and Zhang, Xiangyu and Ren, Shaoqing and Sun, Jian},
  journal={Proceedings of the IEEE conference on computer vision and pattern recognition},
  pages={770--778},
  year={2016}
}
